# Disentangling Alzheimer's disease neurodegeneration from typical brain aging using machine learning

Gyujoon Hwang,[1] Ahmed Abdulkadir,[1] Guray Erus,[1,2] Mohamad Habes,[3] Raymond Pomponio,[1,2] Haochang Shou,[1,4] Jimit Doshi,[1,2] Elizabeth Mamourian,[1,2] Tanweer Rashid,[1,2] Murat Bilgel,[5] Yong Fan,[1,2] Aristeidis Sotiras,[1,6] Dhivya Srinivasan,[1,2] John C. Morris,[7] Daniel Marcus,[6] Marilyn S. Albert,[8] Nick R. Bryan,[9] Susan M. Resnick,[5] Ilya M. Nasrallah,[1,2] Christos Davatzikos,[1,2] David A. Wolk[1,10]; from the iSTAGING consortium; for the ADNI

# Abstract

Neuroimaging biomarkers that distinguish between changes due to typical brain aging and Alzheimer's disease are valuable for determining how much each contributes to cognitive decline. Supervised machine learning models can derive multi-variate patterns of brain change related to the two processes (including the SPARE-AD [Spatial Patterns of Atrophy for Recognition of Alzheimer's Disease] and SPARE-BA [of Brain Aging] scores investigated herein). However, the substantial overlap between brain regions affected in the two processes confounds measuring them independently. We present a methodology, and associated results, toward disentangling the two.

T1-weighted MRI images of 4,054 participants (48–95 years) with Alzheimer's disease, mild cognitive impairment, or cognitively normal diagnoses from the iSTAGING (Imaging-based coordinate SysTem for AGIng and NeurodeGenerative diseases) consortium were analyzed. Multiple sets of SPARE scores were investigated, in order to probe imaging signatures of certain clinically- or molecularly-defined sub-cohorts. First, a subset of clinical Alzheimer's disease patients ($n = 718$) and age- and sex-matched cognitively normal adults ($n = 718$) were selected based purely on clinical diagnoses to train SPARE-BA1 (regression of age using cognitively normal individuals) and SPARE-AD1 (classification of cognitively normal versus Alzheimer's disease) models. Second, analogous groups were selected based on clinical and molecular markers to train SPARE-BA2 and SPARE-AD2 models: amyloid-positive Alzheimer's disease continuum group ($n = 718$; consisting of amyloid-positive Alzheimer's disease, amyloid-positive mild cognitive impairment, and amyloid- and tau-positive cognitively normal individuals) and amyloid-negative cognitively normal group ($n = 718$).

Finally, the combined group of the Alzheimer's disease continuum and amyloid-negative cognitively normal individuals was used to train SPARE-BA3 model, with the intention to estimate brain age regardless of Alzheimer's disease-related brain changes.

The disentangled SPARE models, SPARE-AD2 and SPARE-BA3, derived brain patterns that were more specific to the two types of the brain changes. The correlation between the SPARE-BA Gap (SPARE-BA minus chronological age) and SPARE-AD was significantly reduced after the decoupling ($r$ = 0.56 to 0.06). The correlation of disentangled SPARE-AD was non-inferior to amyloid- and tau-related measurements and to the number of APOE ε4 alleles, but was less to Alzheimer's disease-related psychometric test scores, suggesting contribution of advanced brain aging to these scores. The disentangled SPARE-BA was consistently less correlated with Alzheimer's disease-related clinical, molecular, and genetic variables.

By employing conservative molecular diagnoses and introducing Alzheimer's disease continuum cases to the SPARE-BA model training, we achieved more dissociable neuroanatomical biomarkers of typical brain aging and Alzheimer's disease.

**Author affiliations:**

1 Center for Biomedical Image Computing and Analytics, University of Pennsylvania, Philadelphia, USA

2 Department of Radiology, University of Pennsylvania, Philadelphia, USA

3 Glenn Biggs Institute for Alzheimer's & Neurodegenerative Diseases, University of Texas Health Science Center at San Antonio, San Antonio, USA

4 Penn Statistics in Imaging and Visualization Center, Department of Biostatistics, Epidemiology, and Informatics, Perelman School of Medicine, University of Pennsylvania, Philadelphia, USA

5 Laboratory of Behavioral Neuroscience, National Institute on Aging, Baltimore, USA

6 Department of Radiology, Washington University in St. Louis, St. Louis, USA

7 Department of Neurology, Washington University in St. Louis, St. Louis, USA

8 Department of Neurology, Johns Hopkins University School of Medicine, USA

9 Department of Diagnostic Medicine, University of Texas, Austin; Austin, USA

10 Department of Neurology and Penn Memory Center, University of Pennsylvania, Philadelphia, USA

Correspondence to:

Gyujoon Hwang, Ph.D.

3700 Hamilton Walk, Philadelphia, PA 19104, USA

gyujoon.hwang@pennmedicine.upenn.edu

David A. Wolk, PhD

3615 Chestnut Street, Philadelphia, PA 19104, USA

david.wolk@uphs.upenn.edu

**Abbreviations:** A+ (A-) = Amyloid Positive (Negative); AD = Alzheimer's Disease; ADNI = Alzheimer's Disease Neuroimaging Initiative; ADAS-Cog = Alzheimer's Disease Assessment Scale-Cognitive subscale; AUC = Area Under the Curve; BIOCARD = Biomarkers of Cognitive Decline Among Normal Individuals; BLSA = Baltimore Longitudinal Study of Aging; CN = Cognitively Normal; iSTAGING = Imaging-based coordinate SysTem for AGing and NeurodeGenerative diseases; MAE = Mean Absolute Error; MMSE = Mini-Mental State Examination; MoCA = Montreal Cognitive Assessment; OASIS = Open Access Series of Imaging Studies; RMSE = Root-Mean-Square Error; ROI = Region Of Interest; SPARE-AD = Spatial Pattern of Atrophy for Recognition of Alzheimer's Disease; SPARE-BA = Spatial Pattern of Atrophy for Recognition of Brain Aging; SUVR = Standard Uptake Value Ratio; SVM = Support Vector Machine; SVR = Support Vector Regression; T+ (T-) = Tau Positive (Negative)

# Introduction

Aging is a complex process which can be broadly defined as progressive loss of physiological integrity or gradual accumulation of deleterious biological changes accompanying loss of function.[1,2] While many methods for developing biomarkers of brain aging have been proposed, markers based on structural MRI (commonly known as the "brain age") show less inter-individual variability and methodological variations of measurements relative to other modalities.[3] Generally, brain age is computed by training a multi-dimensional regression model with structural brain features (region- or voxel-based) to predict chronological ages of healthy individuals.[4] This model then looks for the learned structural patterns in an unseen brain and generates a prediction of age. The difference between the predicted brain age and the actual chronological age, often known as the "brain age gap", "brain age gap estimation (brainAGE)", "brain age delta", or "predicted age difference (PAD)", can be used to assess whether the brain looks appropriate for the chronological age or displays deviance from expectation.[2,3,5]

A similar approach can be employed to find structural patterns for other types of brain changes. For example, a machine learning-based score known as the SPARE-AD (Spatial Pattern of Atrophy for Recognition of Alzheimer's Disease) captures multi-variate brain changes of Alzheimer's disease and has been extensively validated.[6,7] It is computed by training a support vector machine (SVM) classification model[8] to separate between cognitively normal and Alzheimer's disease populations using structural brain features and then by measuring the distance of the data points (where each point represents a brain) away from the separating hyperplane.

Summary scores like the SPARE-AD or SPARE-BA, an analogous score for Brain Aging derived using support vector regression (SVR),[9] provide expressive markers that reflect complex brain changes associated with these conditions. As typical brain aging processes and neurodegeneration due to Alzheimer's disease can both affect an individual to varying degrees and account for their cognitive status, one could consider these measures as two axes on a 2D coordinate system where each axis reflects an aspect of structural brain integrity and where each brain can be represented as a point. For example, someone with a similar amount of Alzheimer's disease pathology may be more or less impaired by the degree to which they experience advanced or normal brain aging in this context. Information on which direction on the coordinate system a point deviates from the norm may also be useful.

One caveat with this approach when it comes to Alzheimer's disease is that many brain regions that are associated with Alzheimer's disease are also associated with typical brain aging, which creates inherent correlation between the two scores. We refer to it here as an "entanglement" of the multivariate brain patterns captured by the SPARE-BA and SPARE-AD models or simply an "entanglement" of the two corresponding axes (Fig. 1). For example, brain volumes of regions such as inferior lateral ventricles or middle temporal gyri correlate both with age and the presence of Alzheimer's disease (Fig. 2-A). Therefore, volume changes in these regions would similarly affect both SPARE-BA and SPARE-AD scores, resulting in bias and correlation of the two scores. Disentangling the two can aid constructing more independent and orthogonalized axes on the coordinate system.

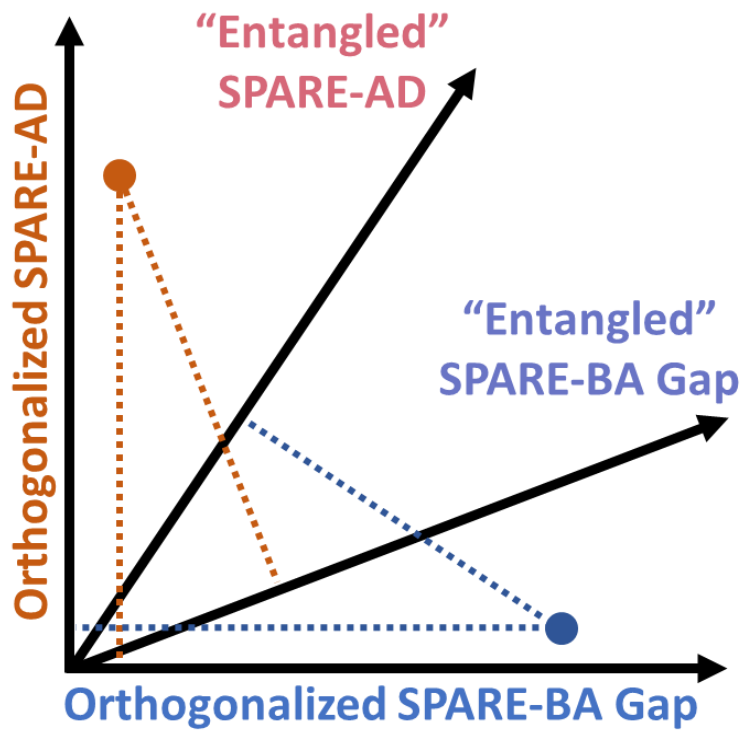

*Figure 1 Hypothetical SPARE scores of two individuals.* *The orange dot on the top left represents a person suffering from Alzheimer's disease, with little advanced brain aging. The blue dot on the bottom right represents a person suffering from advanced brain aging, with little Alzheimer's disease pathology. If the SPARE-AD and SPARE-BA are correlated or "entangled", both individuals would receive elevated SPARE-AD and SPARE BA Gap (SPARE-BA minus chronological age, capturing "advanced" brain aging). The two cases would be better differentiated with orthogonalized or disentangled SPARE scores.*

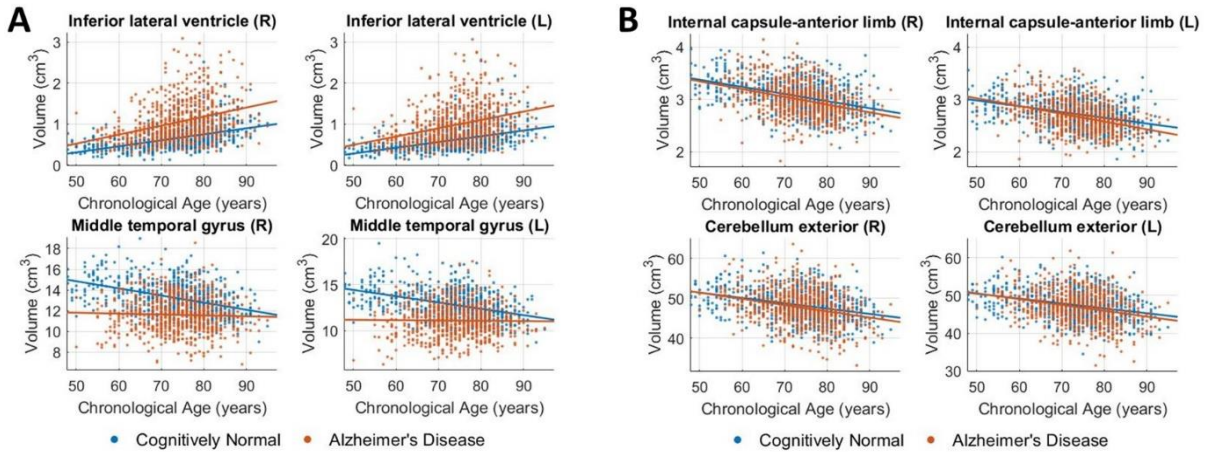

*Figure 2 Examples of brain regions associated with typical aging. Some regions are further influenced by Alzheimer's disease status (A), while others look similar in clinical Alzheimer's disease patients (red) relative to cognitively normal participants (blue) (B). For the SPARE-BA and SPARE-AD scores to be disentangled, we want the SPARE-BA model to be trained with features in (B), which would most likely be ignored by the SPARE-AD model whose objective is to differentiate between the two groups. L = Left hemisphere; R = Right hemisphere.*

Another potential contributor to this unwanted correlation between the two SPARE scores is the low sensitivity and specificity of the diagnosis for Alzheimer's disease.[10] The low sensitivity results in individuals with asymptomatic Alzheimer's disease pathology grouped together with cognitively normal individuals when training the SPARE-BA model. As the proportion of cognitively normal individuals with Alzheimer's disease pathology increases with age, this confound is likely to be more pronounced in the older age range. On the other hand, low specificity results in cognitively impaired individuals with non-Alzheimer's disease pathologies being included in the Alzheimer's disease group, reducing the ability of the SPARE-AD model to detect Alzheimer's disease-specific brain changes. A careful selection of the training samples using molecular biomarkers specific to Alzheimer's disease[11-13] may reduce the correlation between the two SPARE scores.

While Alzheimer's disease and brain aging do display overlap in regions of atrophy that causes unwanted correlation between the two measures, there are reported differences in patterns associated with both processes.[14] Thus, if we can train the SPARE-BA model such that it learns patterns from outside of the Alzheimer's disease-related brain regions (such as regions in Fig.

2-B, instead of in Fig. 2-A), the correlation may be reduced. For example, instead of training the SPARE-BA model using only cognitive normal individuals, which is the usual procedure, we can add individuals with Alzheimer's disease to the training sample. The heterogeneous presence of Alzheimer's disease across the sample would increase variability in Alzheimer's disease-related brain region volumes, reducing strength of their age correlations and making these regions less useful in the age model.

Therefore, the goal of the present study was to derive more orthogonal SPARE-BA and SPARE-AD scores by testing the following two approaches: 1) refine the training samples for both machine learning models by implementing strict molecular diagnostic criteria using amyloid and tau measurements, and 2) add Alzheimer's disease brains in training the SPARE-BA model, which would force the model to learn aging patterns with brain features least affected by Alzheimer's disease. Next, the orthogonalized or disentangled SPARE scores were evaluated based on their correlations with demographic, clinical, molecular, and genetic variables related to Alzheimer's disease.

# Materials and methods

## iSTAGING consortium

iSTAGING (Imaging-based coordinate SysTem for AGing and NeurodeGenerative diseases) is a collection of multi-modal neuroimaging data from more than ten major studies, initiated with the goal of finding reliable and generalizable coordinate systems to capture human brain heterogeneity across a wide range of the age spectrum.[15] Individuals from four studies in the iSTAGING consortium with amyloid status data were investigated: the Alzheimer's Disease Neuroimaging Initiative (ADNI; $n = 1,767$),[16,17] Biomarkers of Cognitive Decline Among Normal Individuals (BIOCARD; $n = 279$),[18] the Baltimore Longitudinal Study of Aging (BLSA; $n = 980$),[19] and the Open Access Series of Imaging Studies (OASIS; $n = 1,028$).[20,21] Only participants 48 years or older were included, as the youngest participant with Alzheimer's disease was 48 years old. Only participants who were classified as cognitively normal, mild cognitive impairment (MCI), or Alzheimer's disease from the originating studies were included in the current analysis. Clinical criteria were broadly similar across the studies and described in Supplementary Table 1. Most participants (3,071 out of 4,054) had data from multiple visits and multiple MRI scans ($n$ = 15,533 total images). If there was a mismatch in acquisition dates between scans and either clinical (18% of cases) or molecular (52% of cases) variables, the search span for date matching was gradually increased (±7, ±30, ±180, ±365 days) until there was a match. If more than one match was found within a range, the measurements were averaged.

The supervisory committee of each study approved its inclusion in this analysis, and this project was approved by the institutional review board of the University of Pennsylvania. All participants gave written informed consent to each study for data acquisition and analyses according to the Declaration of Helsinki. More detail on each of the four studies can be found in Supplementary Methods 1.

## Image preprocessing and harmonization

A fully automated processing pipeline was applied to each T1-weighted image. It involved correction of magnetic field intensity inhomogeneity[22] and multi-atlas skull-stripping using study-specific atlases.[23] Then, 145 brain regions of interest (ROIs) consisting of grey matter, white matter, and ventricular CSF were extracted for each image with a multi-atlas non-linear region segmentation method known as MUSE (MUlti-atlas region Segmentation utilizing Ensembles), which has been validated for its robustness to inter-scanner variabilities.[24,25] Names of the ROIs are listed in Supplementary Table 2.

Significant study-wise effects are inevitable when pooling a diverse dataset for analyses due to lack of standardization in image acquisition protocols and scanner hardware and software.[26] Differences in sample demographics, as well as batch effects, also confound the analyses. Therefore, a systematic quality control and harmonization of the ROIs were performed at the consortium level.[27] This harmonization modeled ROI volumes as a nonlinear function of age and sex using cognitively normal individuals only, and adjusted them for only study-wise differences in mean and variance, thereby preserving age and sex variabilities intrinsic to the data (sex differences are corrected for later as part of the model training). Then the correction coefficients learned from the cognitively normal individuals were applied toward the rest of the dataset. This procedure is based on the method described in detail by Pomponio et al.[27] and has been validated in another work.[15]

# Research design and sample selection

Three different versions of SPARE-BA and SPARE-AD models were designed and were differentiated by suffixes, with each version designed to offer SPARE scores that are more disentangled from the counterpart compared to the preceding version. All versions of SPARE-BA (-BA1, -BA2, -BA3) scores were derived using single linear support vector regression (SVR) models with the only differences being the training samples. A SPARE-BA score is essentially an age predicted by the SVR model using structural brain features (same as "brain age" in the literature).[5] Likewise, SPARE-AD1 and SPARE-AD2 scores were derived using single linear SVM classification models, differing only in the training samples.[7] No two images from a single participant entered the same machine learning model as training samples.

For the first versions of the SPARE scores (SPARE-AD1 and SPARE-BA1), participants were screened with clinical diagnostic criteria obtained from the originating studies. The groups included participants who were labeled as having Alzheimer's disease ("Clinical AD" group; $n = 718$), and the control group included cognitively normal participants ("CN" group; $n = 718$), based solely on their clinical symptoms. Therefore, the SPARE-BA1 regression model was trained with the CN group, and the SPARE-AD1 classification model was trained to separate between the CN and Clinical AD groups.

For the second versions of the SPARE scores (SPARE-AD2 and SPARE-BA2), participants were screened with conservative molecular cutoffs using amyloid and tau measures. CSF β-amyloid 1-42 (Aβ42; provided by ADNI and BIOCARD),[28,29] Pittsburgh compound B ([$^{11}$C]PiB; provided by ADNI, BLSA, and OASIS),[30-32] and [$^{18}$F]florbetapir (also known as [$^{18}$F]AV45; provided by ADNI and OASIS)[33] were used as measures of amyloid, and CSF total tau measurement (provided by ADNI and BIOCARD)[28] was used for tau-related neurodegeneration. CSF total phosphorylated tau (pTau; provided by ADNI and BIOCARD)[34] was also available but was preserved for validation of the results. Conservative cutoffs were used to better assure "positive" and "negative" status, and the methods to define them for each measure per study are detailed in Supplementary Methods 2. The "AD Continuum" group included participants with amyloid positive (A+) results, regardless of their clinical diagnoses: cognitively normal, mild cognitive impairment, or Alzheimer's disease. However, the A+ cognitively normal participants were further required to have an elevated CSF total tau (T+) to increase the likelihood that these individuals were displaying downstream pathology of Alzheimer's disease and likely neurodegeneration. Therefore, the AD Continuum group ($n = $ 718 in total) consisted of A+ Alzheimer's disease ($n = 290$), A+ mild cognitive impairment ($n = 387$) and A+T+ cognitively normal ($n = 41$) individuals. The control group included cognitively normal participants who were amyloid negative ("A-/CN" group; $n = 718$). Therefore, the SPARE-BA2 model (SVR) was trained with the A-/CN group, and the SPARE-AD2 model (SVM) was trained to separate between the A-/CN and AD Continuum groups. To minimize biases, the four groups – CN, Clinical AD (for training SPARE-BA1 and -AD1 models), A-/CN and AD Continuum (for training SPARE-BA2, -BA3, -AD2, and -AD3) – were statistically matched for the sample size, mean age, and sex ratio ($P > 0.2$).

During the sample selection, if multiple scans were available per participant, the last scan with the CN label (or A-/CN), or the first scan with the Clinical AD label (or AD Continuum) was selected. This procedure was done 1) to achieve better age matching between the control and the affected groups and 2) to tighten the SVM separating boundaries by rendering more difficult classification tasks (introducing more borderline cases that serve as "support vectors").[8]

SPARE-BA3 model (SVR) was trained with the combined group of A-/CN and AD Continuum ($n = 1,436$), with the notion that it would avoid Alzheimer's disease-related brain changes. Then the SPARE-AD3 score was computed by regressing out SPARE-BA3 score from SPARE-AD2 score to further reduce the influence of brain aging on the latter. A flowchart for the entire sample selection process is found in Supplementary Fig. 1. The resulting four groups and the demographics are summarized in Table 1.

**Table 1 Demographic summary of all samples**

| Groups | By Clinical Diagnosis | | By Molecular Diagnosis | |
|---|---|---|---|---|
| | Clinical AD ($n$ = 718) | CN ($n$ = 718) | AD Continuum ($n$ = 718) | A-/CN ($n$ = 718) |
| ***n*** | | | | |
| Clinical Diagnosis (AD/MCI/CN) | 718/0/0 | 0/0/718 | 290/387/41 | 0/0/718 |
| Molecular Diagnosis (A+/A-/unclear/no data) | 359/14/11/334 | 47/124/101/446 | 718/0/0/0 | 0/718/0/0 |
| Study (ADNI/BIOCARD/BLSA/OASIS) | 476/1/17/224 | 195/135/194/194 | 621/18/15/64 | 250/100/118/250 |
| **Demographics** | | | | |
| Age (years) [mean ± SD/range] | 74.1 ± 7.6 / 50 – 95 | 73.5 ± 9.3 / 48 – 95 | 73.6 ± 7.6 / 48 – 94 | 73.0 ± 8.8 / 49 – 95 |
| Sex (male/female) | 389 / 329 | 367 / 351 | 389 / 329 | 367 / 351 |
| **Clinical** | | | | |
| MMSE [mean ± SD ($n$)] | 23.1 ± 3.5 (619) | 29.0 ± 1.2 (582) | 25.8 ± 3.5 (649) | 29.0 ± 1.3 (470) |
| **Amyloid** | | | | |
| CSF β-amyloid 42[a] [range ($n$)] | 78 – 275 (128) | 37 – 268 (98) | 37 – 178[b] (238) | 200[b] – 305 (173) |
| [$^{11}$C]PiB SUVR[a] [range ($n$)] | 0.71 – 5.42 (77) | 0.92 – 3.97 (79) | 1.54[b] – 5.42 (96) | 0.50 – 1.25[b] (288) |
| [$^{18}$F]Florbetapir SUVR [range ($n$)] | 0.81 – 3.41 (219) | 0.71 – 2.67 (118) | 1.15[b] – 3.41 (427) | 0.55 – 1.05[b] (310) |

[a]These measurements were harmonized (see Supplementary Methods 1)
[b]These numbers were thresholded.
SD = Standard Deviation

## Model training and testing

To train the SPARE-BA models, the 145 harmonized ROI volumes were first linearly corrected for sex differences and then normalized to *z*-scores. SPARE-BA scores were computed for the training samples using a 10-fold cross-validation, while fine-tuning the gamma parameter. A linear correction of predicted SPARE-BA scores was performed per fold within the training sample to remove known systematic bias caused by regression dilution and regression towards the mean (old individuals predicted young, and vice versa).[4] The model fit was assessed using root-mean-square error (*RMSE*), mean absolute error (*MAE*) and $R^2$.

To train the SPARE-AD1 and SPARE-AD2 models, the harmonized ROI volumes of the control group (CN in SPARE-AD1 and A-/CN in SPARE-AD2) were linearly corrected for sex differences and then normalized to *z*-scores. Then, the correction and scaling factors were transferred to normalize the affected group (Clinical AD in SPARE-AD1 and AD Continuum in SPARE-AD2). SPARE-AD scores were computed for the training samples using a 10-fold cross-validation, while fine-tuning the *C* regularization parameter. *C* which yielded equal sensitivity and specificity was selected for the final model. The model fit was assessed using the area-under-the-curve (*AUC*) and classification accuracy at separating between the control and the affected groups. The SPARE-BA3 scores were linearly regressed out from the SPARE-AD2 scores using a 10-fold cross-validation for creation of SPARE-AD3.

All ten cross-validated models per version of a SPARE model including all correction and normalization factors were saved and applied to the testing samples. The resulting ten values per a SPARE score were averaged for the final score. If a testing sample was from a participant that entered the model training, then only one of the ten models in which the participant was left out was applied.

## Orthogonalization of SPARE scores

Whether the SPARE-BA and SPARE-AD scores were disentangled was assessed in three steps. First, the correlation between the SPARE-BA Gap (SPARE-BA minus chronological age) and SPARE-AD was computed using the scores from all participants ($n = 4{,}054$). If multiple images were available per participant, then only one was selected randomly, while favoring images that were labeled clinically as Alzheimer's disease over mild cognitive impairment over cognitively normal. Decreased correlation of these two SPARE scores was considered as evidence of orthogonalization. Second, the ability of the SPARE-BA Gap and SPARE-AD to separate between CN and Clinical AD and between A-/CN and AD Continuum was tested. We hypothesized that the orthogonalization would significantly decrease this separability using SPARE-BA Gap as it would be less sensitive to Alzheimer's disease-related changes, which was the desired goal. SPARE score cutoffs that yielded equal sensitivity and specificity were used to compute the classification accuracy.

Third, the weights assigned to the 145 brain ROIs by each machine learning model were evaluated and compared between models to assess how much of the patterns found by the SPARE-BA and SPARE-AD models overlapped. To examine the statistical significance of the weights, a permutation test was performed by retraining each version of the model 5,000 times with the labels shuffled (chronological ages in the SVR models, and the group memberships in the SVM models). The null distribution of the weights was plotted per ROI and compared with the observed weights to derive two-tailed $P$-values.[35]

## Clinical, molecular, and genetic associations

The correlations between the SPARE scores and the molecular measures were computed. In addition to the amyloid and tau measures used for diagnosis, CSF total pTau was tested. 48% of the SPARE scores had molecular measures from the same day, and the date difference of the rest was 69.3±64.3 days with the maximum of 365 days. Since there existed large study-wise differences in measurements, only one study with the largest sample was analyzed per variable. To examine more localized tau pathology, tau PET measurements, available only in a subset of participants in ADNI,[36] from entorhinal area and inferior temporal gyrus were also correlated with the SPARE scores.

We also computed correlations between the SPARE scores and clinical test scores related to Alzheimer's disease. These tests included Mini-Mental State Examination (MMSE),[37] Montreal Cognitive Assessment (MoCA),[38] Alzheimer's Disease Assessment Scale-Cognitive Subscale (ADAS-Cog),[39,40] Logical Memory,[41] and Trail Making Test.[42] To examine whether the disentangled SPARE-BA contributed to the variance in these scores in patients beyond SPARE-AD, nested multivariate linear regression models were trained, first with SPARE-BA3, SPARE-BA3 Gap, and SPARE-AD3 as predictors, and subsequently with excluding one predictor at a time. The correlations between the SPARE scores and the number of APOE ε4 alleles[43] were also computed.

## Statistical analysis

Two-sample *t*-tests were used to examine if the mean age was matched between the four groups – CN, Clinical AD, A-/CN and AD Continuum (raw $P > 0.2$). Chi-squared tests were used to examine if the sex ratio was matched between the groups (raw $P > 0.2$).

Pearson correlation was used to compute correlations between the SPARE scores. Spearman correlation was used to compute correlations between the SPARE scores and clinical, molecular, or genetic variables, as some variables were non-normally distributed (e.g., MMSE score with the cap at 30) or discretized (e.g., number of APOE ε4 alleles). When checking for changes in correlation between SPARE-BA Gap and SPARE-AD, the significance of difference between the two coefficients was determined with the *F*-statistic computed on Fisher *z*-transformed coefficients. When comparing between two nested models, the likelihood ratio test[44] was used to assess the goodness of fit.

## Data availability

Raw MRI images as well as all demographic, clinical, and genetic information used in this analysis can be obtained from databases of each of the four studies with a reasonable request upon approval. Population-based results from the iSTAGING consortium including the SPARE models are projected to be available by the end of 2021.

# Results

## Molecular diagnosis

1,810 out of 3,257 (56%) images from individuals designated as cognitively normal at the time of the scan and with amyloid measurements passed our conservative amyloid cutoffs to be included in the A-/CN group. 688 out of 814 (85%) images from individuals designated as Alzheimer's disease at the time of scan and with amyloid measurements, 926 out of 1,798 (52%) images from individuals labeled as mild cognitive impairment, and 78 out of 3,257 (2%) images labeled as cognitively normal filled our amyloid and tau criteria to be included in the AD Continuum group. After selecting a single image per participant as well as statistically matching for mean age, sex ratio, and the number of unique participants, the A-/CN and AD

Continuum groups included 718 participants in each, whose demographics are summarized in Table 1. These sample sizes guided the selection of participants into the final CN and Clinical AD groups ($n = 718$ in each) for the training of SPARE-BA1 and SPARE-AD1 models. The mean age and sex ratio were statistically matched between all four groups ($P > 0.2$).

## Orthogonalization of SPARE scores

Machine learning cross-validation results of all versions of SPARE models are summarized in Supplementary Table 3. Overall, SPARE-BA2 showed the best fit ($MAE = 4.92$, $R^2 = 0.645$) among the three versions of SPARE-BA, as the training sample was the most homogeneous. When computed on a common test dataset, age correlation of the SPARE-BA3 was better than that of the SPARE-BA1, both for cognitively normal individuals ($MAE = 5.80$ to $6.20$, $n = 1,218$) and for individuals with either mild cognitive impairment or Alzheimer's disease ($MAE = 5.90$ to $8.23$, $n = 506$) (Supplementary Fig. 2). The SPARE-AD1 model showed better cross-validation results ($AUC = 0.89$) than SPARE-AD2 ($AUC = 0.82$) as the task of separating between CN and Clinical AD groups was relatively easier compared to separating between A-/CN and AD Continuum.

However, note that the goal here was not to achieve best fit results, but to build models that produce disentangled SPARE scores. The correlation between SPARE-BA Gap and SPARE-AD was the highest between SPARE-BA1 and SPARE-AD1 ($r = 0.553$, $n = 4,054$), which was significantly reduced between SPARE-BA2 and SPARE-AD2 ($r = 0.406$, difference $P < 0.001$), and again between SPARE-BA3 and SPARE-AD2 ($r = 0.062$, difference $P < 0.001$) (Fig. 3). In cognitively normal individuals, SPARE-AD3 was also significantly less correlated with chronological age ($r = 0.124$, $n = 2,388$) compared to SPARE-AD2 ($r = 0.237$, difference $P < 0.001$), whose correlation was less compared to SPARE-AD1 ($r = 0.332$, difference $P < 0.001$).

The classification accuracy of the SPARE-BA Gap at separating between A-/CN and AD Continuum was gradually reduced from 67.4% with SPARE-BA1 to 63.6% with SPARE-BA2, then to 55.9% with SPARE-BA3, suggesting progressively greater specificity to age and non-Alzheimer's disease effects. Alternatively, SPARE-AD remained robust at this classification among the three versions (> 74%). Similarly, the classification accuracy of SPARE-BA Gap at separating between CN and Clinical AD was reduced from 75.4% to 70.9%, then to 58.6%, while that of SPARE-AD remained relatively robust (> 80%) (Fig. 4). These results support the notion that SPARE-BA3 is not conflated with either biomarker-driven classification or clinical diagnosis of Alzheimer's disease.

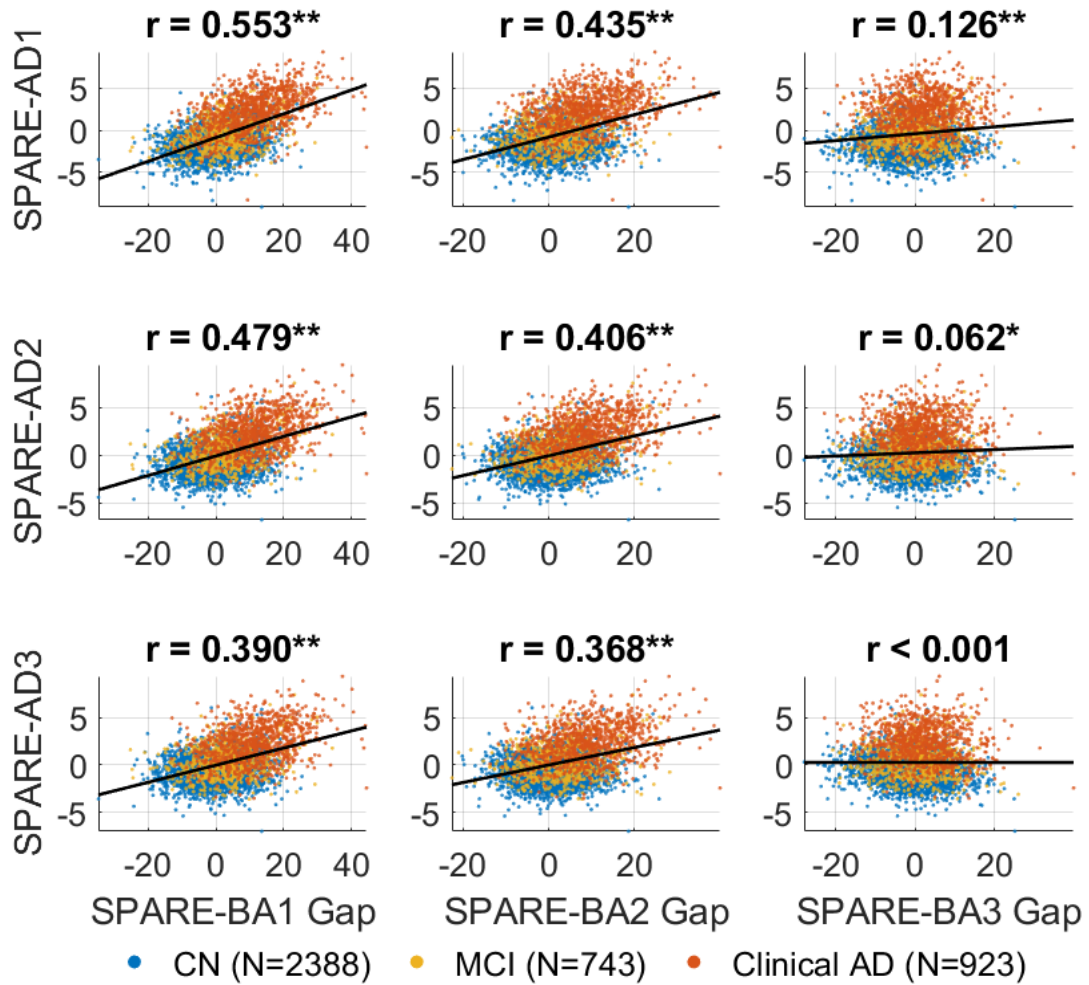

*Figure 3 Correlations between SPARE-BA Gap and SPARE-AD.* *Decreased correlation between SPARE-BA Gap (SPARE-BA minus chronological age) and SPARE-AD was considered as evidence of orthogonalization. Pearson correlation coefficients including all data points (n = 4,054) are labeled on top of each subplot. CN = Cognitively Normal; MCI = Mild Cognitive Impairment; AD = Alzheimer's Disease; \*P < 0.05; \*\*P < 0.0001*

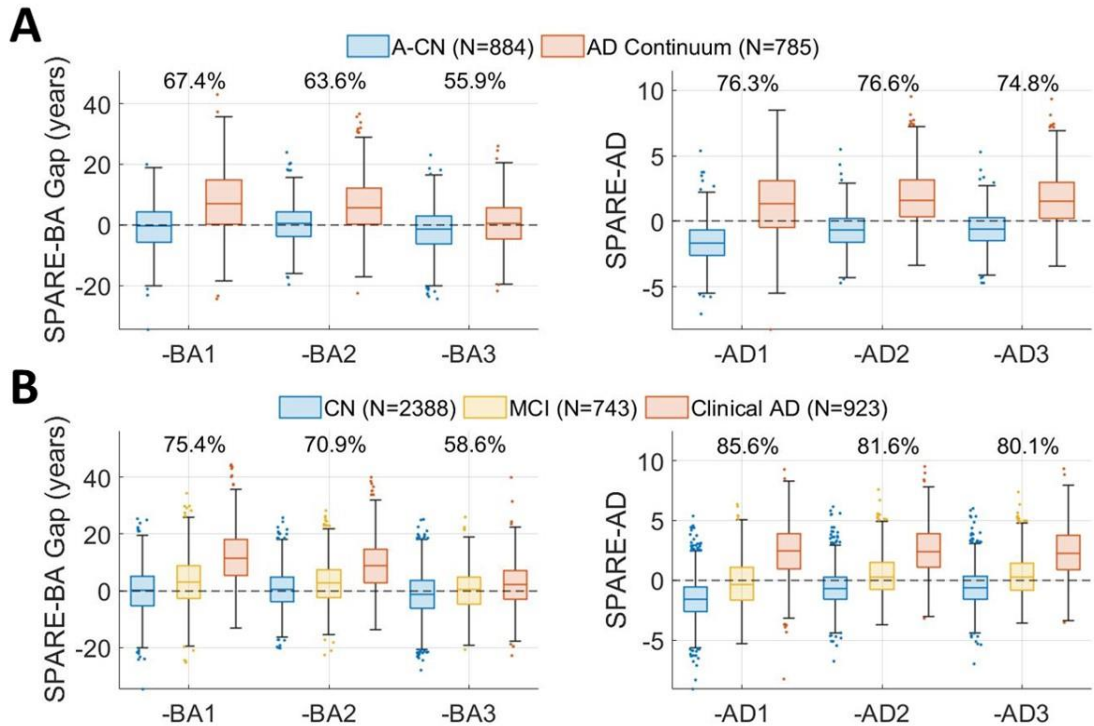

*Figure 4 Distributions of SPARE scores by diagnoses.* *Numbers above the boxes are the classification accuracies at (**A**) separating between A-CN and AD Continuum (combined group of A+ Alzheimer's disease, A+ mild cognitive impairment, and A+T+ cognitively normal) and (**B**) separating between CN and Clinical AD, where sensitivity equals specificity. CN = Cognitively Normal; MCI = Mild Cognitive Impairment; AD = Alzheimer's Disease.*

## Disentangled regional atrophy patterns

The brain ROIs that received most significant weights or that showed greatest changes in weights between the versions of the SPARE machine learning models were evaluated (Table 2). Left thalamus, right planum temporale, and bilateral anterior limb of internal capsule received significant weights in all SPARE-BA models ($P < 0.02$, permutation analysis), but not in any SPARE-AD models ($P > 0.4$). Regions including bilateral cerebellum exterior, right parahippocampal gyrus, and postcentral gyrus received more significant weights in SPARE-BA3 ($P < 0.03$), compared to SPARE-BA1 and SPARE-BA2 ($P > 0.07$).

On the other hand, left hippocampus, left lingual gyrus and left postcentral gyrus received significant weights in both SPARE-AD1 and SPARE-AD2 models ($P < 0.04$), but not in any SPARE-BA models ($P > 0.1$). Regions including right inferior lateral ventricle and bilateral superior occipital gyri received less significant weights in SPARE-BA3 ($P > 0.17$), compared to SPARE-BA1 and SPARE-BA2 ($P < 0.05$).

**Table 2 ROIs with the most significant model weights or with the greatest changes in weights**

| ROI Name | SPARE-BA1 | SPARE-BA2 | SPARE-BA3 | SPARE-AD1 | SPARE-AD2 |
|---|---|---|---|---|---|
| **Only Significant in SPARE-BA** | | | | | |
| Thalamus Proper (L) | **-3.43(<0.001)** | **-3.84(<0.001)** | **-3.52(0.005)** | 0.05(0.852) | -0.18(0.488) |
| Anterior Limb of Internal Capsule (R) | **-3.45(<0.001)** | **-4.03(<0.001)** | **-5.55(<0.001)** | 0.14(0.551) | 0.14(0.576) |
| Anterior Limb of Internal Capsule (L) | **-2.48(0.005)** | **-3.45(0.001)** | **-5.99(<0.001)** | 0.09(0.664) | -0.04(0.865) |
| Planum Temporale (R) | **-3.23(0.013)** | **-3.65(0.006)** | **-3.90(0.009)** | 0.05(0.671) | -0.02(0.884) |
| **Become More Significant in SPARE-BA / Become Less Significant in SPARE-AD** | | | | | |
| Caudate (R) | 1.71(0.101) | **3.83(<0.001)** | **4.91(<0.001)** | 0.02(0.919) | -0.07(0.754) |
| Cerebellum Exterior (R) | -1.80(0.078) | -1.49(0.165) | **-3.19(0.006)** | -0.30(0.233) | 0.17(0.531) |
| Cerebellum Exterior (L) | -1.50(0.136) | -1.20(0.274) | **-3.24(0.006)** | 0.47(0.056) | 0.14(0.594) |
| Orbital Part of the Inferior Frontal Gyrus (R) | -1.44(0.327) | **-3.19(0.036)** | **-3.62(0.012)** | -0.08(0.331) | -0.01(0.863) |
| Parahippocampal Gyrus (R) | 1.77(0.156) | 2.06(0.111) | **3.23(0.023)** | -0.02(0.882) | 0.07(0.621) |
| Postcentral Gyrus (R) | 0.70(0.571) | -2.22(0.098) | **-3.83(0.011)** | 0.21(0.073) | 0.16(0.211) |
| Brain Stem | **4.51(<0.001)** | **3.21(0.003)** | **6.31(<0.001)** | **0.36(0.033)** | 0.14(0.428) |
| Planum Polare (R) | **-4.29(0.001)** | **-2.86(0.032)** | **-4.49(0.003)** | **0.27(0.029)** | 0.12(0.340) |
| Posterior Limb of Internal Capsule (L) | 0.76(0.463) | 0.18(0.859) | 1.11(0.404) | **-0.44(0.016)** | 0.12(0.543) |
| Middle Frontal Gyrus (L) | 1.23(0.277) | 0.06(0.960) | 2.59(0.069) | **-0.43(0.004)** | -0.12(0.490) |
| Middle Temporal Gyrus (R) | -1.68(0.148) | -1.10(0.388) | -0.36(0.797) | **-0.44(0.006)** | 0.00(0.985) |
| Precuneus (L) | 0.04(0.971) | -0.43(0.716) | 2.52(0.067) | **-0.38(0.021)** | -0.12(0.491) |
| Supplementary Motor Cortex (R) | -0.81(0.502) | -1.96(0.138) | -0.11(0.930) | **-0.31(0.013)** | -0.15(0.267) |
| **Only Significant in SPARE-AD** | | | | | |
| Hippocampus (L) | -1.23(0.280) | -1.80(0.127) | -0.77(0.552) | **-0.51(0.014)** | **-0.60(0.005)** |
| Lingual Gyrus (L) | -1.33(0.287) | 0.29(0.820) | -0.43(0.768) | **0.38(0.002)** | **0.28(0.038)** |
| Postcentral Gyrus Medial Segment (L) | -0.15(0.915) | 1.25(0.405) | 1.23(0.392) | **-0.23(0.012)** | **-0.27(0.004)** |
| **Become More Significant in SPARE-AD / Become Less Significant in SPARE-BA** | | | | | |
| Inferior Lateral Ventricle (R) | **4.40(<0.001)** | **6.15(<0.001)** | 0.80(0.547) | 0.19(0.245) | **0.49(0.004)** |
| Medial Orbital Gyrus (R) | 0.80(0.520) | 1.43(0.279) | 0.74(0.619) | 0.14(0.288) | **0.30(0.018)** |
| Occipital Fusiform Gyrus (L) | 0.35(0.804) | -0.68(0.639) | -2.23(0.130) | 0.13(0.192) | **0.34(0.001)** |
| Parietal Operculum (L) | -0.05(0.970) | 0.80(0.556) | 1.19(0.415) | 0.15(0.226) | **0.38(0.002)** |
| Supramarginal Gyrus (L) | **-2.94(0.029)** | -0.69(0.626) | 0.29(0.840) | -0.06(0.562) | **-0.24(0.032)** |
| Fornix (L) | **-4.32(0.001)** | -1.33(0.343) | -1.01(0.487) | 0.17(0.102) | 0.14(0.189) |
| Superior Frontal Gyrus Medial Segment (L) | **-3.45(0.005)** | -1.39(0.291) | -1.95(0.175) | 0.21(0.113) | -0.07(0.600) |
| Posterior Insula (L) | **-3.58(0.003)** | 0.01(0.997) | 0.15(0.927) | 0.04(0.784) | -0.03(0.839) |
| Superior Occipital Gyrus (R) | **-2.99(0.027)** | **-2.81(0.043)** | -1.99(0.172) | -0.20(0.057) | -0.20(0.077) |
| Superior Occipital Gyrus (L) | **-2.94(0.021)** | **-3.32(0.013)** | -0.81(0.586) | 0.11(0.311) | 0.04(0.742) |

Numbers in parentheses are two-tailed *P*-values that were derived from permutation tests. Weights with *P*<0.05 are highlighted.
A positive weight indicates that the model associated the increase in the ROI volume to the positive case (older age in SPARE-BA models, and AD in SPARE-AD models), while ROIs with insignificant weights were less informative to the models.
L = left hemisphere; R = right hemisphere.

# Clinical, molecular, and genetic associations

Consistent with greater specificity to aging, SPARE-BA3 did not significantly correlate with any tau measures in individuals with either mild cognitive impairment or Alzheimer's disease. Further, it was significantly less correlated with amyloid and tau measures, compared to SPARE-BA2 and SPARE-BA1 (Table 3 and Supplementary Table 4). SPARE-AD2 correlated more strongly with Aβ42 ($\rho$ = -0.412, $n$ = 773) and [$^{18}$F]florbetapir ($\rho$ = 0.422, $n$ = 759) compared to SPARE-AD1.

**Table 3 Spearman correlations between SPARE scores and Alzheimer's disease-related variables**

| Variable | $n$ | Age | SPARE-BA1 | SPARE-BA2 | SPARE-BA3 | SPARE-AD1 | SPARE-AD2 | SPARE-AD3 |
|---|---|---|---|---|---|---|---|---|
| **Molecular (amyloid)** | | | | | | | | |
| CSF Aβ42 | 773 | -0.095* | -0.256** | -0.239** | -0.071† | -0.392** | **-0.412**** | -0.409** |
| [$^{18}$F]Florbetapir (AV45) SUVR | 759 | 0.104* | 0.265** | 0.239** | 0.094*† | 0.403** | **0.422**** | 0.418** |
| **Molecular (tau)** | | | | | | | | |
| CSF Total Tau | 760 | 0.132* | 0.254** | 0.217** | 0.059† | **0.427**** | 0.404** | 0.402** |
| CSF Total Phosphorylated Tau | 773 | 0.018 | 0.124* | 0.100* | -0.040† | 0.302** | 0.306** | **0.310**** |
| Tau PET (entorhinal area) | 258 | 0.008 | 0.242* | 0.192* | 0.035† | **0.477**** | 0.430** | 0.428** |
| Tau PET (inferior temporal gyrus) | 258 | -0.069 | 0.199* | 0.144* | -0.007 | 0.402** | 0.410** | **0.415**** |
| **Psychometric** | | | | | | | | |
| MMSE | 1310 | -0.170** | -0.500** | -0.462** | -0.257**† | **-0.600**** | -0.554** | -0.537** |
| MOCA | 836 | -0.257** | -0.537** | -0.499** | -0.309**† | **-0.583**** | -0.539** | -0.519** |
| ADAS-Cog 13 | 1306 | 0.188** | 0.537** | 0.493** | 0.263**† | **0.669**** | 0.619**† | 0.602** |
| Logical Memory (delayed) | 1244 | -0.130** | -0.458** | -0.416** | -0.225**† | **-0.611**** | -0.549**† | -0.536** |
| Trail Making Test (part A) | 1215 | 0.178** | **0.423**** | 0.407** | 0.248**† | 0.394** | 0.361** | 0.342** |
| **Genetic** | | | | | | | | |
| APOE4 Alleles‡ | 1753 | -0.123** | 0.108** | 0.093** | -0.027*† | 0.293** | 0.293** | **0.300**** |

Correlations in individuals with either mild cognitive impairment or Alzheimer's disease are shown. More comprehensive table can be found in Supplementary Table 3. Highest correlations per variable are highlighted.
*Corrected $P$ < 0.05; **Corrected $P$ < 0.0001
†Significant difference from the value on the left ($P$ < 0.05).
‡Both patients and cognitively normal individuals

SPARE-AD2 and SPARE-AD3 ($\rho$ = 0.310, $n$ = 773) were more correlated with CSF total phosphorylated tau compared to SPARE-AD1 ($\rho$ = 0.302), while the opposite pattern was observed for CSF total tau ($\rho$ = 0.427 with SPARE-AD1 and $\rho$ = 0.402 with SPARE-AD3, $n$ = 760). SPARE-AD1 was the strongest predictor of tau PET standard uptake value ratio (SUVR) in entorhinal area ($\rho$ = 0.477, $n$ = 258), whereas SPARE-AD3 was the strongest predictor of tau PET SUVR in inferior temporal gyrus ($\rho$ = 0.415, $n$ = 258). When considering patients and cognitively normal individuals together, SPARE-AD3 was more correlated with the number of APOE ε4 alleles ($\rho$ = 0.300, $n$ = 1,753) compared to SPARE-AD2 and SPARE-AD1 ($\rho$ = 0.293) (Table 3). Nonetheless, differences in correlation between SPARE-AD scores and molecular measures and APOE ε4 were not statistically significant, supporting the notion that the changes in approach, including regressing out brain age in SPARE-AD3, did not diminish its specificity to Alzheimer's disease. In the context of SPARE-BA3 not displaying correlation with these molecular measures, these findings support our success in orthogonalization to the underlying biology of Alzheimer's disease.

SPARE-BA3 was also significantly less correlated with all Alzheimer's disease-related clinical test scores examined herein compared to SPARE-BA2 and SPARE-BA1 (difference $P < 0.05$) (Table 3). Nonetheless, its correlations with all scores remained significant, and either SPARE-BA3 or SPARE-BA3 Gap was still a significant contributor to the multivariate models predicting the scores according to the likelihood ratio test (Supplementary Table 5), suggesting that it provided additional explanatory value beyond SPARE-AD3 even when decoupled from Alzheimer's disease-related brain changes. This supports the notion that brain aging itself contributes to varying degrees to the cognitive phenotype of patients with Alzheimer's disease.

# Discussion

Developing reliable and expressive metrics that summarize complex, high-dimensional brain patterns is critical in neuroimaging. The SPARE scores are designed to capture patterns of specific brain changes to serve as summary metrics. However, given overlap in affected brain regions, structural brain-based measures of typical brain aging and Alzheimer's disease are expected to correlate. Here, we show that this correlation can be effectively eliminated by carefully restricting the training samples and by introducing abnormal cases to the brain aging models.

## Molecular definition of Alzheimer's disease

With the advancements in both biofluid and imaging-based biomarkers, the field has been moving towards a more biologically-based definition of Alzheimer's disease.[12,45] This has recently been formulated in the National Institute of Aging-Alzheimer's Association (NIA-AA) research framework,[46] in which patients are dichotomously classified along three dimensions: amyloid (A), tau (T), and neurodegeneration (N).[47] Parallel to neuropathological definitions of Alzheimer's disease, one must have the presence of both amyloid and tau biomarkers (A+, T+) to be classified as having Alzheimer's disease. These efforts reflect the fact that diagnosis of Alzheimer's disease based solely on cognitive symptoms is not sufficient and may cause significant false positive and false negative results.[10]

In this context, we tested whether defining training samples based on biological markers such as amyloid and tau can improve the specificity of the structural brain imaging markers. As any measurement would inherently contain variance, we excluded individuals at the borderline by using stricter cutoffs than what are commonly used. Only 1,810 out of 3,257 (56%) images from individuals who were labeled as cognitively normal passed our conservative cutoff for being A-, while 688 out of 814 (85%) images from individuals that were labeled clinically as Alzheimer's disease passed the cutoff for being A+. This supports the heterogeneity that exists in samples that are solely defined by clinical symptoms, and as discussed earlier, misclassified individuals result in measures of brain aging and Alzheimer's disease that are tainted by the other. The resulting model after employing molecular cutoffs, SPARE-AD2, assigned higher weights to the temporal lobe regions compared to the SPARE-AD1 model. SPARE-AD2 and SPARE-AD3 scores were generally more correlated with amyloid measures and similarly with measures of CSF total phosphorylated tau (pTau) and tau PET. Importantly, SPARE-AD2 and SPARE-AD3 were significantly less correlated with chronological age, which supported their increased specificity to Alzheimer's disease-related brain changes. Moreover, using these molecular definitions also modestly reduced the sensitivity of SPARE-BA2 to differentiating Alzheimer's disease versus cognitively normal individuals.

## Specific brain biomarkers of brain aging and Alzheimer's disease

Our results showed that by defining the training samples based on molecular markers, the undesired correlation between SPARE-BA and SPARE-AD could be significantly reduced. By adding affected individuals (AD Continuum group) to the brain aging model, the correlation was further reduced, indeed almost eliminated. The resulting machine learning models learned brain patterns that were more specific to each of the two types of brain changes. For example, the SPARE-BA3 model avoided brain regions such as the inferior lateral ventricle, whose volume increase was associated not only with brain aging in general, but also with the presence of Alzheimer's disease. This was also evidenced by the progressively reduced relationship of the SPARE-BA scores (from SPARE-BA1 to SPARE-BA3) with Alzheimer's disease-related molecular markers.

Conflating biomarkers of typical brain aging and Alzheimer's disease results in biased assessment of a person's brain. However, whether an individual with Alzheimer's disease has an otherwise healthy brain or whether there is significant neurodegeneration related to non-Alzheimer's disease causes is important information for clinicians to estimate prognosis and consider whether comorbidity-directed interventions may be worthwhile. Using the disentangled SPARE scores, one may be able to better differentiate between individuals with Alzheimer's disease but otherwise healthy brain (top-left quadrant in the scatter plots in Fig. 3) and individuals with accelerated brain aging but no Alzheimer's disease (bottom-right quadrant in the scatter plots in Fig. 3). This can potentially assist in determining the degree to which Alzheimer's disease is driving neurodegeneration including the relative contribution of both processes to cognition and other outcomes in any individual, which may have implications for treatment and intervention studies.

## Correlation with clinical variables

SPARE-AD2 and SPARE-AD3 were similarly or better correlated with amyloid and tau related measurements, as well as with the number of APOE4 alleles, compared to SPARE-AD1. This suggests that the disentangled SPARE-AD, in addition to being less conflated with age, is still robustly linked to Alzheimer's disease pathology. However, their correlations with psychometric test scores were somewhat reduced after this disentangling compared to SPARE-AD1. This is likely because these cognitive variables are not as specific to the Alzheimer's disease pathology alone, but also associated with the degree of background brain aging[48] or other non-AD neurodegenerative processes which are often mixed with Alzheimer's disease. Therefore, SPARE-AD1, which is sensitive to not only Alzheimer's disease-related brain changes, but also to advanced brain aging, likely better tracks with these mixed cognitive changes. Indeed, SPARE-AD1 is trained based on clinical status, which is closely linked to psychometric performance, whereas SPARE-AD2 and AD3 are trained based on molecular status regardless of the performance, including cognitively normal individuals. Given these results, the disentangled SPARE-AD scores may not always be ideal if trying to track overall clinical status, especially where cognitive impairment reflects the effects of a number of pathologies, and different versions of SPARE scores may be used in combination to construct a more comprehensive description of an individual's brain.

The correlations between SPARE-BA3 and Alzheimer's disease-related clinical, molecular, and genetic variables were consistently reduced and non-significant for the latter two, compared to SPARE-BA1 or SPARE-BA2. This emphasizes that the disentangled SPARE-BA scores are no longer influenced by brain changes associated specifically with Alzheimer's disease, which would be useful for assessing the presence of accelerated brain aging in individuals with Alzheimer's disease without bias. Moreover, despite being disentangled, SPARE-BA3 Gap still significantly correlated with the psychometric measures, further supporting the notion that accelerated aging itself is associated with declines in cognition, or, alternatively, that decelerated brain aging is associated with resilience to age-associated cognitive decline. Thus, the combination of SPARE-BA3 with more selective SPARE-AD scores (SPARE-AD2 and SPARE-AD3), may best capture the contributions of both processes to cognitive decline, as evidenced by the complementary inclusion in nested multivariate models of cognition.

## Sample selection

The results underscore the importance of sample selection in building brain-based prediction models such as the SPARE models. Inclusion and exclusion criteria used to define training samples significantly influenced neurodegenerative patterns learned by the models as well as the relationships of the derived scores to other variables. Therefore, caution is needed when the SPARE or other similar summary scores are used in novel populations or clinical settings to fully understand the implications of the scores based upon the methods applied to develop the underlying models. Here, amyloid status was primarily applied to select samples for the disentangled SPARE models, firstly, because of its high specificity to Alzheimer's disease, but also because amyloid was the most commonly available molecular marker in the iSTAGING database. Tau status, or the combination of imaging-based molecular markers specific to Alzheimer's disease can further improve the SPARE models to be highly focused.

## Comorbidities of aging

Aging is a complex process that accompanies many comorbidities, including cancer, diabetes, cardiovascular disease, and neurodegenerative disease.[1] Differentiating between typical aging and aging-related disorders is difficult, if at all possible.[49,50] The meaning of what is "typical" and the degree to which a variety of age-associated processes contribute to brain integrity remain unclear. However, one may conceive that the current state of a human brain might be decomposed into the normative state corresponding to the chronological age plus weighted sum of abnormalities which further modulate brain structure. In this setting, the status of the brain can be described by a feature space where each axis corresponds to a type of an abnormality. While this is an oversimplification, as there are likely to be second order interactions and non-linearity of combined effects. However, developing such a simplified dimensional space may help researchers and clinicians by providing a deeper understanding of brain health.

Here, we introduced approaches to measure two such dimensions: advanced brain aging and Alzheimer's disease. There are many more axes to be added to this feature space, with each addition also potentially refining the normative population sample as well. For example, in the current analysis, the normal group (A-/CN) was defined as individuals without amyloid pathology. But it could, for example, further exclude those with diabetes from the control group and evaluate diabetes status distinctly as was done for Alzheimer's disease here. A SPARE-diabetes score can be disentangled from SPARE-BA and SPARE-AD. As such, in the future, one brain scan may produce numerous such SPARE scores that are disentangled from each other, assessing the state of a human brain by measuring the presence and severity of neurodegeneration related to many pathologies.

## Limitations and future work

There are a few limitations to the current work. First, there is variability in molecular measures across studies and techniques. Even within a single imaging technique, such as florbetapir PET, there exists a variety of methods for preparation and processing of the results, which causes significant between-study differences. Here, we circumvented the issue by linearly harmonizing the diagnostic cutoffs to match sensitivity and specificity, but the analyses linking the measures to the SPARE scores had to be performed in a study-wise manner.

Second, intuitively, a regression model is more suitable for developing biomarkers such as the SPARE scores than a classification model. This is because the separating plane in a classification model is most heavily governed by the borderline cases (especially with SVM). For example, in the SPARE-AD2 model, the separating plane was most likely defined by the more advanced cases of A-/CN and the least severe cases of AD Continuum in terms of the brain changes. This may explain the lower-than-expected classification accuracy of SPARE-AD2 at separating between the two groups and may have limited the correlation between SPARE-AD2 and Alzheimer's disease-related molecular measurements. Here, a classification model was selected for SPARE-AD to be consistent with previous studies, but future work can explore regression model-based SPARE-AD scores.

Despite the limitations, the current work successfully demonstrated a decoupling of the SPARE-BA and SPARE-AD. Future work can explore tau status in combination with amyloid to more carefully define Alzheimer's disease as more tau data become available. Additionally, the methods used here may be expanded and validated using more advanced, non-linear machine learning models as they have been shown in other systems to improve overall performance.[51,52] Also, additional efforts to disentangle SPARE-BA from other aging-related comorbidities may be valuable in delineating a complete set of most objective and specific biomarkers to evaluate an aging brain.

# Acknowledgements

The authors would like to thank the investigators of ADNI, BIOCARD, BLSA, and OASIS, as well as all the participants and their families. For further acknowledgements, please see the supplementary material.

# Funding

This work was supported, in part, by NIH grants 1RF1-AG054409-01, U01-AG068057, and P30-AG010124. ADNI is supported by NIH grants U01-AG024904 and RC2-AG036535. The Preclinical Alzheimer's disease Consortium (PAC) includes BLSA, BIOCARD, and OASIS and is supported by NIH grant RF1-AG059869. The BIOCARD study is supported by NIH grant U19-AG033655. The BLSA is supported by the Intramural Research Program, National Institute on Aging, and Research and Development Contract HHSN-260-2004-00012C. The Adult Children's Study (ACS), which provides its data through OASIS, is supported by NIH grant P01-AG026276.

# Competing interests

Dr. Nasrallah was an educational speaker for Biogen. Dr. Wolk has received grant support from Merck, Biogen, and Eli Lilly/Avid and consultation fees from Neuronix and GE Healthcare and is on the Data and Safety Monitoring Board for a clinical trial run by Functional Neuromodulation.

# Supplementary Material

**Supplementary Table 1. Summary of diagnostic criteria from pooled studies**

| Study | Diagnosis | Criteria Summary | Reference |
|---|---|---|---|
| ADNI | CN, MCI, AD | Memory complaint by patient or study partner Abnormal memory function score on Logical Memory II subscale (delayed paragraph recall, paragraph A only) from the Wechsler Memory Scale – revised **MCI**: Mini-Mental State Exam (MMSE) score between 24 and 30, CDR = 0.5, Memory Box score at least 0.5 **AD:** MMSE score between 20 and 26 (AD), CDR = 0.5 or 1.0, Memory Box score at least 1.0 | http://adni.loni.usc.edu/methods/documents/ |
| BIOCARD | CN, MCI, AD | For individuals with a CDR > 0 and/or evidence of decline on cognitive testing, we followed the clinical diagnostic criteria in the National Institute on Aging/Alzheimer's Association (NIA/AA) working group reports for the diagnosis of **MCI**[1] and dementia due to **AD**[2] | Albert et al.[1] Sacktor et al.[3] |
| BLSA | CN, MCI, AD | For individuals whose Blessed Information Memory Concentration score[4] was 3 or greater, CDR > 0, or Dementia Questionnaire[5] was abnormal, **AD** diagnosis was based on Diagnostic and Statistical Manual of Mental Disorders, Revised Third Edition criteria,[6] and **MCI** diagnosis was based on the Mayo Clinic criteria.[7] Accordingly, the diagnosis of MCI was given to subjects who had deficits limited to one or two areas of cognition (usually memory), with preservation of normal activities of daily living compared with other people of similar age. | O'Brien et al.[8] |
| OASIS | CN, AD | The diagnosis was based on clinical information (derived primarily from a collateral source). **AD**: CDR > 0 | Marcus et al.[9] |

AD: Alzheimer's disease, MCI: Mild Cognitive Impairment, CDR: Clinical Dementia Rating, CN: Cognitively Normal

**Supplementary Methods 1. Summary of Pooled Studies**

Data used in the preparation of this article were obtained in part from the Alzheimer's Disease Neuroimaging Initiative (ADNI) database (adni.loni.usc.edu).[10,11] The ADNI was launched in 2003 as a public-private partnership, led by Principal Investigator Michael W. Weiner, MD. The primary goal of ADNI has been to test whether serial MRI, PET, other biological markers, and clinical and neuropsychological assessment can be combined to measure the progression of mild cognitive impairment (MCI) and early Alzheimer's disease. For up-to-date information, see www.adni-info.org.

The BIOCARD Study (officially entitled "Biomarkers of Cognitive Decline Among Normal Individuals: the BIOCARD cohort") is an extension of the Family Study/BIOCARD Study begun at National Institute of Mental Health (NIMH) in 1995.[12] The overarching goal of the BIOCARD Study is to identify biomarkers associated with progression from normal cognitive status to cognitive impairment or dementia, with a particular focus on Alzheimer's disease. Investigators at the Johns Hopkins University School of Medicine began evaluating participants in 2009. The subjects are seen annually. At each visit there are assessments of medical and cognitive status, as well as acquisition of MRI, CSF, PET-PiB, and blood. The Johns Hopkins research team is the coordinating site of the Preclinical Alzheimer's disease Consortium (PAC).

The Baltimore Longitudinal Study of Aging (BLSA) is an ongoing longitudinal study led by the National Institute of Aging (NIA).[13] Established in Baltimore, Maryland, the USA in 1958, BLSA aims to characterize the process of aging in general, and healthy aging in particular. Beginning in 1986, BLSA introduced an extensive neuropsychological assessment to subjects older than 60 years. The neuroimaging sub-study of BLSA began in 1994, with annual or semi-annual MRI studies, cognitive testing, and clinical evaluations of a subset of 721 BLSA participants. For up-to-date information, see www.blsa.nih.gov.

The Open Access Series of Imaging Studies (OASIS) is a series of neuroimaging data sets that is publicly available for study and analysis.[9,14] It is aimed at making neuroimaging datasets freely available to the scientific community and facilitating future discoveries in basic and clinical neuroscience. Previously released data for OASIS-Cross-sectional and OASIS-Longitudinal have been utilized for hypothesis driven data analyses, development of neuroanatomical atlases, and development of segmentation algorithms. OASIS-3 is a longitudinal neuroimaging, clinical, cognitive, and biomarker dataset for normal aging and Alzheimer's disease. The OASIS datasets hosted by central.xnat.org provide the community with open access to a significant database of neuroimaging and processed imaging data across a broad demographic, cognitive, and genetic spectrum an easily accessible platform for use in neuroimaging, clinical, and cognitive research on normal aging and cognitive decline. For up-to-date information, see www.oasis-brains.org.

**Supplementary Table 2. 145 brain ROIs that were used as training features**

| | |
|---|---|
| 3rd ventricle | (R/L) Gyrus rectus |
| 4th ventricle | (R/L) Inferior occipital gyrus |
| (R/L) Accumbens area | (R/L) Inferior temporal gyrus |
| (R/L) Amygdala | (R/L) Lingual gyrus |
| Brain Stem | (R/L) Lateral orbital gyrus |
| (R/L) Caudate | (R/L) Middle cingulate gyrus |
| (R/L) Cerebellum exterior | (R/L) Medial frontal cortex |
| (R/L) Cerebellum (white matter) | (R/L) Middle frontal gyrus |
| (R/L) Hippocampus | (R/L) Middle occipital gyrus |
| (R/L) Inferior lateral ventricle | (R/L) Medial orbital gyrus |
| (R/L) Lateral ventricle | (R/L) Postcentral gyrus medial segment |
| (R/L) Pallidum | (R/L) Precentral gyrus medial segment |
| (R/L) Putamen | (R/L) Superior frontal gyrus medial segment |
| (R/L) Thalamus proper | (R/L) Middle temporal gyrus |
| (R/L) Ventral diencephalon | (R/L) Occipital pole |
| Cerebellar vermal lobules I-V | (R/L) Occipital fusiform gyrus |
| Cerebellar vermal lobules VI-VII | (R/L) Opercular part of inferior frontal gyrus |
| Cerebellar vermal lobules VIII-X | (R/L) Orbital part of inferior frontal gyrus |
| (R/L) Basal forebrain | (R/L) Posterior cingulate gyrus |
| (R/L) Frontal lobe (white matter) | (R/L) Precuneus |
| (R/L) Occipital lobe (white matter) | (R/L) Parahippocampal gyrus |
| (R/L) Parietal lobe (white matter) | (R/L) Posterior insula |
| (R/L) Temporal lobe (white matter) | (R/L) Parietal operculum |
| (R/L) Fornix | (R/L) Postcentral gyrus |
| (R/L) Anterior limb of internal capsule | (R/L) Posterior orbital gyrus |
| (R/L) Posterior limb of internal capsule including cerebral peduncle | (R/L) Planum polare |
| Corpus callosum | (R/L) Precentral gyrus |
| (R/L) Anterior cingulate gyrus | (R/L) Planum temporale |
| (R/L) Anterior insula | (R/L) Subcallosal area |
| (R/L) Anterior orbital gyrus | (R/L) Superior frontal gyrus |
| (R/L) Angular gyrus | (R/L) Supplementary motor cortex |
| (R/L) Calcarine cortex | (R/L) Supramarginal gyrus |
| (R/L) Central operculum | (R/L) Superior occipital gyrus |
| (R/L) Cuneus | (R/L) Superior parietal lobule |
| (R/L) Entorhinal area | (R/L) Superior temporal gyrus |
| (R/L) Frontal operculum | (R/L) Temporal pole |
| (R/L) Frontal pole | (R/L) Triangular part of the inferior frontal gyrus |
| (R/L) Fusiform gyrus | (R/L) Transverse temporal gyrus |

R/L: Right and Left hemisphere

**Supplementary Methods 2. Amyloid and tau cutoffs**

Conservative amyloid and tau cutoffs were selected to refine the training samples in the second and third versions of the SPARE models (SPARE-BA2, -BA3, -AD2, and -AD3). CSF β-amyloid 1-42 (Aβ42) measures were provided by ADNI and BIOCARD, but, given differences in acquisition methods, there were significant study-wise discrepancies. First, for the ADNI participants, Aβ42<180pg/mL was labeled as amyloid positive (A+) and Aβ42>200pg/mL was labeled as amyloid negative (A-). Participants with Aβ42 between the two values were excluded from the training. Note that based on prior work, Aβ42<192pg/mL has been considered consistent with presence of cerebral amyloid using the Luminex platform.[15] Then, the equivalent cutoffs for Aβ42 measurements from BIOCARD[16] were calculated by matching two normal distribution fits: one for cognitively normal participants and another for individuals with either mild cognitive impairment or Alzheimer's disease diagnosed clinically (MCI/AD). The equivalent cutoffs were 354pg/mL and 391pg/mL from comparing the distributions of the first groups, and 350pg/mL and 391pg/mL from comparing the distributions of the second groups. Then, the final cutoffs for BIOCARD were the average of the two sets (Aβ42<352pg/mL for A+ and Aβ42>391pg/mL for A-). Sensitivity (percentage of A+ in MCI/AD group) and specificity (percentage of A- in cognitively normal group) of Aβ42 were 70% and 55% for ADNI, and 61% and 57% for BIOCARD.

Pittsburgh compound B ([$^{11}$C]PiB) standardized uptake value ratio (SUVR) measurements from amyloid-PET were provided by ADNI, BLSA, and OASIS, but there was study-wise discrepancy in measurements. First, for OASIS, [$^{11}$C]PiB>1.50 (high amyloid burden group according to their previous study) was labeled as A+ and [$^{11}$C]PiB<1.25 (low burden group) was labeled as A-.[17] Using similar technique as with Aβ42 (only changing MCI/AD to only Alzheimer's disease as OASIS did not have participants with mild cognitive impairment), the harmonized cutoffs for ADNI were 1.59 and 1.66 (a cutoff of 1.6 has been used in their previous study)[18]; for BLSA they were 1.01 and 1.09 (a cutoff of 1.064 has been used in their previous study).[19] Sensitivity (percentage of A+ in Alzheimer's disease group) and specificity of [$^{11}$C]PiB were 79% and 50% for ADNI, 67% and 65% for BLSA, 73% and 74% for OASIS.

[$^{18}$F]florbetapir SUVR measurements from amyloid-PET were provided by ADNI and OASIS and their distributions were similar in terms of the cutoffs. Previously, a cutoff of 1.11 was established.[20] To be conservative, [$^{18}$F]florbetapir>1.15 was labeled as A+ and [$^{18}$F]florbetapir<1.05 was labeled as A- for both studies. Sensitivity (percentage of A+ in

Alzheimer's disease group) and specificity of [$^{18}$F]florbetapir were 84% and 51% for ADNI, and 90% and 56% for OASIS. If more than one of the above three amyloid measurements were available per visit, all labels needed to agree for the final label.

Finally, CSF total tau measurements were provided by ADNI and BIOCARD and their distributions were similar in terms of the cutoffs. CSF total tau was only used to screen A+ cognitively normal participants to be included in the AD Continuum group. Previously, a cutoff of 93pg/mL was validated.[15] To be conservative, CSF total tau>100pg/mL was labeled as tau positive (T+) for both studies. 17% of cognitively normal participants in ADNI and 12% of those in BIOCARD were labeled as T+.

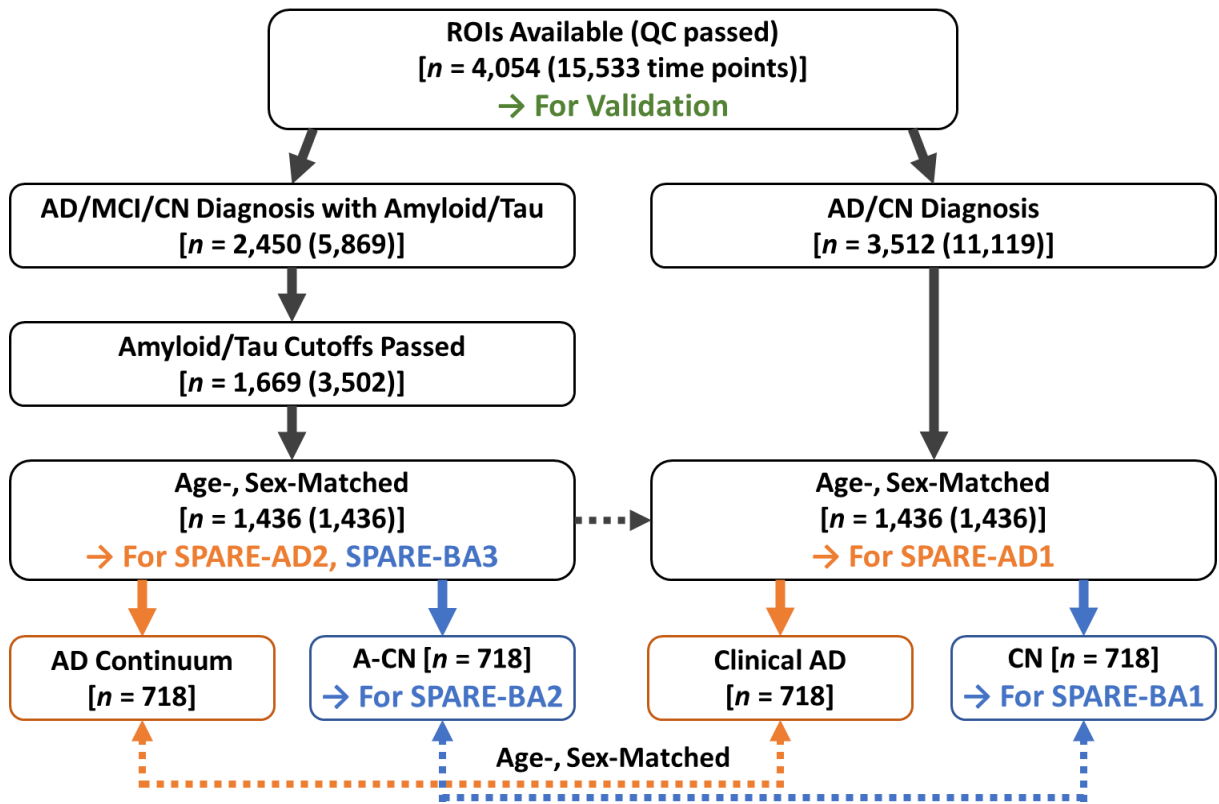

**Supplementary Figure 1 A flowchart of the entire sample selection process.** Age- and sex-matched "AD Continuum" and "A-CN" groups were selected first, and then the "Clinical AD" and "CN" groups were selected to match between the four groups. *n* indicates the number of unique participants, while the numbers in parentheses indicate the numbers of total images. No two images from a participant entered a single model as training samples. AD = Alzheimer's Disease; CN = Cognitively Normal; MCI = Mild Cognitive Impairment; QC = Quality Control

**Supplementary Table 3 10-fold cross-validation results from the SPARE model training**

| SVR Results | MAE/RMSE/$R^2$ | SVM Results | AUC / Accuracy ± SD (%) |
|---|---|---|---|
| SPARE-BA1 | 6.16/6.14/0.566 | SPARE-AD1 | 0.89 / 83.6 ± 2.9 |
| SPARE-BA2 | 4.92/5.22/0.645 | SPARE-AD2 | 0.82 / 74.4 ± 3.4 |
| SPARE-BA3 | 5.94/5.73/0.513 | SPARE-AD3[a] | - |

[a]SPARE-AD3 was not an SVM model, but rather a simple model to residualize SPARE-AD2 with SPARE-BA3 scores.
AUC = Area-Under-the-Curve; MAE = Mean Absolute Error; RMSE = Root-Mean-Squared Error; SD = Standard Deviation.

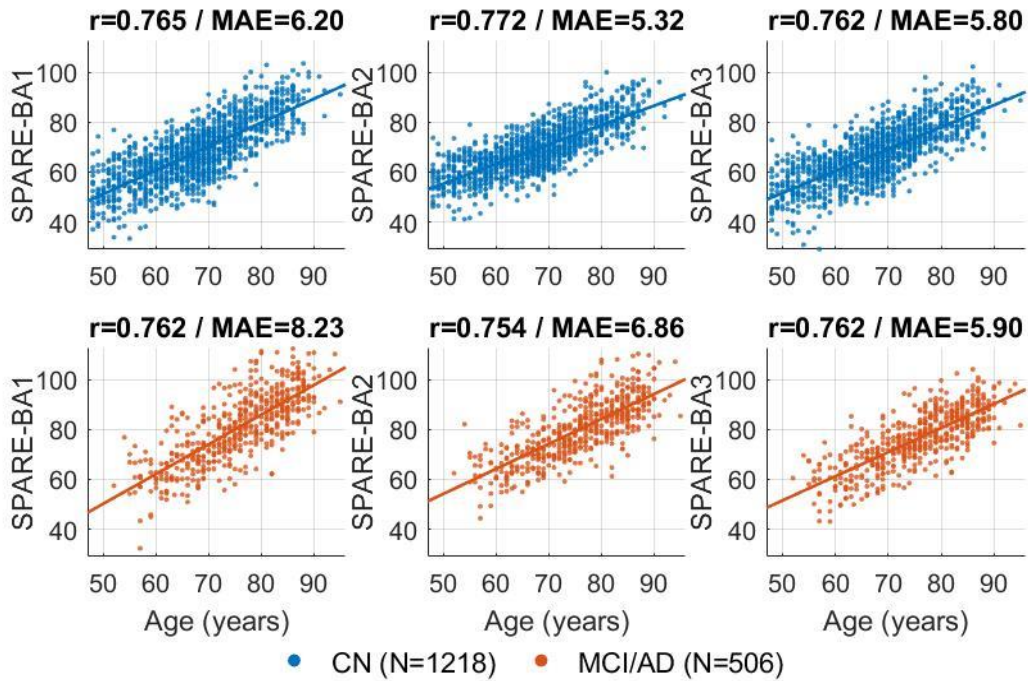

**Supplementary Figure 2 SPARE-BA test results.** Correlation between SPARE-BA and chronological age was computed on individuals whose images were not used in the training. The expected fit line is y = x. The linear fit is relatively similar in all three versions in cognitively normal (CN) individuals, while it is improved in individuals with either mild cognitive impairment (MCI) or Alzheimer's disease (AD). CN = Cognitively Normal; MCI/AD = Mild Cognitive Impairment/Alzheimer's Disease; MAE = Mean Absolute Error

# Supplementary Table 4 Spearman correlations between SPARE scores and Alzheimer's disease-related variables

| Variable | Study | Group | n | AGE | SPARE-BA1 | SPARE-BA2 | SPARE-BA3 | SPARE-BA1 Gap | SPARE-BA2 Gap | SPARE-BA3 Gap | SPARE-AD1 | SPARE-AD2 | SPARE-AD3 |
|---|---|---|---|---|---|---|---|---|---|---|---|---|---|
| **Molecular (amyloid)** | | | | | | | | | | | | | |
| CSF Aβ42 | ADNI | ALL | 1055 | -0.126** | -0.309** | -0.296** | -0.129**† | -0.291** | -0.262** | -0.065*† | -0.433** | **-0.452**** | -0.444** |
| | ADNI | CN | 282 | -0.220** | -0.138* | -0.164* | -0.149* | 0.002 | 0.007 | -0.016 | -0.051 | -0.121 | -0.098 |
| | ADNI | MCI+AD | 773 | -0.095* | -0.256** | -0.239** | -0.071† | -0.255** | -0.220** | -0.009† | -0.392** | -0.412** | -0.409** |
| [18F]Florbetapir (AV45) SUVR | ADNI | ALL | 1091 | 0.081* | 0.250** | 0.229** | 0.088*† | 0.256** | 0.218** | 0.049† | 0.405** | **0.428**** | 0.424** |
| | ADNI | CN | 332 | 0.005 | -0.016 | -0.006 | -0.033 | -0.033 | -0.029 | -0.047 | 0.095 | 0.177* | 0.178* |
| | ADNI | MCI+AD | 759 | 0.104* | 0.265** | 0.239** | 0.094*† | 0.263** | 0.208** | 0.029† | 0.403** | 0.422** | 0.418** |
| [11C]PiB SUVR | OASIS | ALL | 484 | **0.337**** | 0.329** | 0.316** | 0.269** | 0.082 | -0.024 | -0.022 | 0.226** | 0.188** | 0.153* |
| | OASIS | CN | 436 | 0.279** | 0.215** | 0.205** | 0.185* | -0.029 | -0.116* | -0.073 | 0.085 | 0.055 | 0.024 |
| | ADNI | MCI+AD | 83 | -0.110 | 0.009 | -0.019 | -0.081 | 0.153 | 0.083 | 0.035 | 0.236* | 0.211 | 0.216 |
| **Molecular (tau)** | | | | | | | | | | | | | |
| CSF Total Tau | ADNI | ALL | 1041 | 0.161** | 0.288** | 0.263** | 0.105*† | 0.233** | 0.186** | -0.010† | **0.430**** | 0.424** | 0.418** |
| | ADNI | CN | 281 | 0.237** | 0.098 | 0.140* | 0.115 | -0.081 | -0.066 | -0.051 | -0.006 | 0.087 | 0.076 |
| | ADNI | MCI+AD | 760 | 0.132* | 0.254** | 0.217** | 0.059† | 0.217** | 0.158** | -0.055† | 0.427** | 0.404** | 0.402** |
| CSF Total Phosphorylated Tau | ADNI | ALL | 1054 | 0.056 | 0.153** | 0.133** | -0.001† | 0.141** | 0.115* | -0.042† | 0.310** | 0.321** | **0.321**** |
| | ADNI | CN | 281 | 0.154* | -0.049 | -0.032 | -0.023 | -0.197* | -0.222* | -0.166* | -0.027 | 0.049 | 0.050 |
| | ADNI | MCI+AD | 773 | 0.018 | 0.124* | 0.100* | -0.040† | 0.148** | 0.124* | -0.057† | 0.302** | 0.306** | 0.310** |
| Tau PET (entorhinal area) | ADNI | ALL | 635 | 0.104* | 0.232** | 0.199** | 0.109* | 0.202** | 0.127* | 0.051 | 0.377** | 0.285** | 0.274** |
| | ADNI | CN | 377 | 0.054 | 0.036 | 0.014 | 0.016 | -0.003 | -0.038 | -0.022 | 0.132* | 0.032 | 0.028 |
| | ADNI | MCI+AD | 258 | 0.008 | 0.242* | 0.192** | 0.035† | 0.294** | 0.200** | 0.038† | **0.477**** | 0.430** | 0.428** |
| Tau PET (inferior temporal gyrus) | ADNI | ALL | 635 | 0.051 | 0.195** | 0.160* | 0.081 | 0.211** | 0.150* | 0.081 | 0.290** | 0.273** | 0.266** |
| | ADNI | CN | 377 | 0.055 | 0.053 | 0.035 | 0.037 | 0.031 | 0.009 | 0.025 | 0.093 | 0.084 | 0.078 |
| | ADNI | MCI+AD | 258 | -0.069 | 0.199* | 0.144* | -0.007 | 0.322** | 0.233** | 0.073† | 0.402** | 0.410** | **0.415**** |
| **Psychometric** | | | | | | | | | | | | | |
| MMSE | ADNI | ALL | 1764 | -0.156** | -0.511** | -0.478** | -0.273**† | -0.524** | -0.465**† | -0.233**† | **-0.629**** | -0.589** | -0.572** |
| | BLSA | CN | 849 | -0.201** | -0.196** | -0.190** | -0.181** | -0.046 | 0.031 | 0.001 | -0.115* | -0.093* | -0.072* |
| | ADNI | MCI+AD | 1310 | -0.170** | -0.500** | -0.462** | -0.257**† | -0.502** | -0.428**† | -0.192**† | -0.600** | -0.554** | -0.537** |
| MOCA | ADNI | ALL | 1194 | -0.248** | -0.521** | -0.495** | -0.318**† | -0.487** | -0.420**† | -0.194**† | **-0.596**** | -0.546** | -0.523** |
| | ADNI | CN | 358 | -0.307** | -0.260** | -0.276** | -0.234** | -0.101 | -0.070 | -0.037 | -0.235** | -0.143* | -0.110* |
| | ADNI | MCI+AD | 836 | -0.257** | -0.537** | -0.499** | -0.309**† | -0.497** | -0.404**† | -0.153**† | -0.583** | -0.539** | -0.519** |
| ADAS-COG-11 | ADNI | ALL | 1760 | 0.163** | 0.533** | 0.501** | 0.277**† | 0.542** | 0.482**† | 0.225**† | **0.672**** | 0.635** | 0.616** |
| | ADNI | CN | 453 | 0.116* | 0.113* | 0.132* | 0.098* | 0.067 | 0.083 | 0.032 | 0.168* | 0.194* | 0.184* |
| | ADNI | MCI+AD | 1307 | 0.165** | 0.507** | 0.464** | 0.240**† | 0.511** | 0.432**† | 0.168**† | 0.642** | 0.601** | 0.584** |
| ADAS-COG-13 | ADNI | ALL | 1759 | 0.189** | 0.567** | 0.532** | 0.304**† | 0.563** | 0.497**† | 0.232**† | **0.694**** | 0.651**† | 0.631** |
| | ADNI | CN | 453 | 0.202** | 0.196** | 0.203** | 0.174* | 0.096 | 0.093 | 0.049 | 0.163* | 0.182* | 0.161* |
| | ADNI | MCI+AD | 1306 | 0.188** | 0.537** | 0.493** | 0.263**† | 0.527** | 0.446**† | 0.172**† | 0.669** | 0.619** | 0.602** |
| Logical Memory (immediate) | ADNI | ALL | 1681 | -0.087* | -0.491** | -0.455** | -0.244**† | -0.536** | -0.485**† | -0.253**† | **-0.633**** | -0.575**† | -0.560** |
| | ADNI | CN | 437 | 0.006 | -0.079 | -0.078 | -0.066 | -0.086 | -0.089 | -0.080 | -0.046 | -0.095 | -0.087 |
| | ADNI | MCI+AD | 1244 | -0.099* | -0.462** | -0.417** | -0.211**† | -0.490** | -0.421**† | -0.192**† | -0.585** | -0.511**† | -0.498** |
| Logical Memory (delayed) | ADNI | ALL | 1681 | -0.116** | -0.502** | -0.466** | -0.262**† | -0.532** | -0.481**† | -0.250**† | **-0.659**** | -0.609**† | -0.593** |
| | ADNI | CN | 437 | -0.031 | -0.105* | -0.097 | -0.082 | -0.098 | -0.096 | -0.078 | -0.071 | -0.113* | -0.100* |
| | ADNI | MCI+AD | 1244 | -0.130** | -0.458** | -0.416** | -0.225**† | -0.466** | -0.399**† | -0.176**† | -0.611** | -0.549**† | -0.536** |
| Trail Making Test (part A) | ADNI | ALL | 1627 | 0.213** | **0.461**** | 0.446** | 0.287**† | 0.417** | 0.370** | 0.187**† | 0.455** | 0.409** | 0.388** |
| | BLSA | CN | 896 | 0.425** | 0.385** | 0.386** | 0.380** | 0.029 | -0.112*† | -0.035 | 0.181** | 0.150** | 0.096* |
| | ADNI | MCI+AD | 1215 | 0.178** | 0.423** | 0.407** | 0.248**† | 0.396** | 0.348** | 0.156**† | 0.394** | 0.361** | 0.342** |
| Trail Making Test (part B) | ADNI | ALL | 1608 | 0.204** | 0.493** | 0.466** | 0.295**† | 0.468** | 0.410**† | 0.218**† | **0.512**** | 0.474** | 0.451** |
| | BLSA | CN | 900 | 0.386** | 0.356** | 0.370** | 0.359** | 0.035 | -0.079*† | 0.001 | 0.157** | 0.081* | 0.024 |
| | ADNI | MCI+AD | 1195 | 0.186** | 0.466** | 0.435** | 0.265**† | 0.451** | 0.385** | 0.184**† | 0.457** | 0.426** | 0.405** |
| **Genetic** | | | | | | | | | | | | | |
| APOE4 Alleles | ADNI | ALL | 1753 | -0.123** | 0.108** | 0.093** | -0.027† | 0.214** | 0.212** | 0.082*† | 0.293** | 0.293** | **0.300**** |
| | BLSA | CN | 831 | -0.144** | -0.125** | -0.132** | -0.129** | -0.011 | 0.032 | 0.003 | -0.007 | -0.021 | -0.001 |
| | ADNI | MCI+AD | 1301 | -0.144** | 0.060* | 0.045 | -0.060*† | 0.179** | 0.172** | 0.062*† | 0.258** | 0.248** | 0.256** |

Highest correlations per variable are highlighted.
*Corrected $P < 0.05$; **Corrected $P < 0.0001$
†Significant difference from the value on the left ($P < 0.05$).

**Supplementary Table 5 Multivariate linear regression models using orthogonalized SPARE scores to predict clinical variables**

| Response Variable (y) | n | SPARE-BA3 | SPARE-BA3 Gap | SPARE-AD3 | Likelihood Ratio (p)[21]† |
|---|---|---|---|---|---|
| MMSE | 1310 | -0.105* | -0.090* | -0.496** | |
| | | -0.164** | | -0.491** | 0.0047 |
| | | | -0.158** | -0.509** | 0.0018 |
| MOCA | 836 | -0.183** | -0.005 | -0.463** | |
| | | -0.186** | | -0.463** | **0.8987** |
| | | | -0.123** | -0.492** | <0.0001 |
| ADAS-Cog 11 | 1307 | 0.090* | 0.092* | 0.533** | |
| | | 0.150** | | 0.527** | 0.0041 |
| | | | 0.150** | 0.542** | 0.0047 |
| ADAS-Cog 13 | 1306 | 0.115* | 0.090* | 0.548** | |
| | | 0.173** | | 0.543** | 0.0041 |
| | | | 0.163** | 0.562** | 0.0007 |
| Logical Memory (immediate) | 1244 | -0.064 | -0.117* | -0.457** | |
| | | -0.141** | | -0.453** | 0.0010 |
| | | | -0.159** | -0.464** | **0.0570** |
| Logical Memory (delayed) | 1244 | -0.118* | -0.073* | -0.478** | |
| | | -0.166** | | -0.475** | 0.0302 |
| | | | -0.149** | -0.491** | 0.0010 |
| Trail Making Test (part A) | 1215 | 0.072 | 0.096* | 0.291** | |
| | | 0.137** | | 0.287** | 0.0111 |
| | | | 0.144** | 0.297** | **0.0534** |
| Trail Making Test (part B) | 1195 | 0.129* | 0.074* | 0.359** | |
| | | 0.178** | | 0.356** | 0.0426 |
| | | | 0.158** | 0.370** | 0.0010 |

Only individuals with mild cognitive impairment (MCI) or Alzheimer's disease (AD) were included in the models.
y ~ SPARE-BA3 + SPARE-BA3 Gap + SPARE-AD3 for unrestricted models (first row in each y). Blank cells indicate dropped variables.
*Corrected $P < 0.05$; **Corrected $P < 0.0001$
†Significance indicates that the restricted model fit is significantly worse than the unrestricted model (corrected $P > 0.05$ are highlighted)

**Further Acknowledgements**

Data used in the preparation of this article were in part obtained from the Alzheimer's Disease Neuroimaging Initiative (ADNI) database. As such, the investigators within the ADNI contributed to the design and implementation of ADNI and provided data but did not participate in analysis or writing of this report. For a complete list of investigators involved in ADNI see: http://www.loni.ucla.edu/ADNI/Data/ADNI_Authorship_List.pdf.

ADNI is funded by the National Institute on Aging, the National Institute of Biomedical Imaging and Bioengineering, and through generous contributions from the following: Abbott, AstraZeneca AB, Bayer Schering Pharma AG, Bristol-Myers Squibb, Eisai Global Clinical Development, Elan Corporation, Genentech, GE Healthcare, GlaxoSmithKline, Innogenetics, Johnson and Johnson, Eli Lilly and Co., Medpace, Inc., Merck and Co., Inc., Novartis AG, Pfizer Inc, F. Hoffman-La Roche, Schering-Plough, Synarc, Inc., and Wyeth, as well as nonprofit partners the Alzheimer's Association and Alzheimer's Drug Discovery Foundation, with participation from the US Food and Drug Administration. Private sector contributions to ADNI are facilitated by the Foundation for the National Institutes of Health (www.fnih.org). The grantee organization is the Northern California Institute for Research and Education, and the study is coordinated by the Alzheimer's Disease Cooperative Study at the University of California, San Diego. ADNI data are disseminated by the Laboratory for Neuro Imaging at the University of California, Los Angeles. This research was also supported by NIH grants P30 AG010129, K01 AG030514, and the Dana Foundation. The National Cell Repository for Alzheimer's Disease (NIH grant U24 AG021886) provided support for DNA and cell line banking and processing for ADNI.